\pdfoutput=1

\documentclass[11pt]{article}

\usepackage[final]{acl}

\usepackage{times}
\usepackage{latexsym}
\usepackage{fontawesome}
\usepackage{array}

\usepackage{booktabs} 
\usepackage{multirow} 
\usepackage[T1]{fontenc}

\usepackage[utf8]{inputenc}

\usepackage{microtype}

\usepackage{inconsolata}

\usepackage{graphicx}

\title{Injecting Domain-Specific Knowledge into Large Language Models: \\A Comprehensive Survey}

\def\SP{~~}

\author{
\textmd{Zirui Song}$^{1\ast}$
\SP
\textmd{Bin Yan}$^2$\thanks{Equal contributions.}
\SP 
\textmd{Yuhan Liu}$^{3}$
\SP
\textmd{Miao Fang}$^2$
\SP 
\textmd{Mingzhe Li}$^4$ \\
\SP
\textmd{Rui Yan}$^{5}$
\SP
\textmd{Xiuying Chen}$^{1}$\thanks{Corresponding author.}
\\[0.1125cm]
\normalsize 
$ ^1$Mohamed bin Zayed University of Artificial Intelligence 
\normalsize \\
\normalsize
\SP ~~~
$ ^2 $Northeastern University
\SP ~~~
$ ^3 $Gaoling School of Artificial Intelligence, Renmin University of China
\normalsize
\\
\normalsize
\SP ~~~
$ ^4 $ByteDance
\SP ~~~
$ ^5$Wuhan University
\normalsize
\\
\\
{
\normalsize E-mail: \normalsize \tt   zirui.song@mbzuai.ac.ae, xiuying.chen@mbzuai.ac.ae }\\
}

\begin{document}
\maketitle

\begin{abstract}
Large Language Models (LLMs) have demonstrated remarkable success in various tasks such as natural language understanding, text summarization, and machine translation. 
However, their general-purpose nature often limits their effectiveness in domain-specific applications that require specialized knowledge, such as healthcare, chemistry, or legal analysis.
To address this, researchers have explored diverse methods to enhance LLMs by integrating domain-specific knowledge. 
In this survey, we provide a comprehensive overview of these methods, which we categorize into four key approaches: dynamic knowledge injection, static knowledge embedding, modular adapters, and prompt optimization. 
Each approach offers unique mechanisms to equip LLMs with domain expertise, balancing trade-offs between flexibility, scalability, and efficiency. 
We discuss how these methods enable LLMs to tackle specialized tasks, compare their advantages and disadvantages, evaluate domain-specific LLMs against general LLMs, and highlight the challenges and opportunities in this emerging field. 
For those interested in delving deeper into this area, we also summarize the commonly used datasets and benchmarks.
To keep researchers updated on the latest studies, we maintain an open-source at: \faGithub ~\href{https://github.com/abilliyb/Knowledge_Injection_Survey_Papers}{\textcolor{blue}{official-repo.com}}, dedicated to documenting research in the field of specialized LLM.

\end{abstract}

\section{Introduction}
\begin{figure*}[tb]
    \centering
    \includegraphics[width=0.95\linewidth]{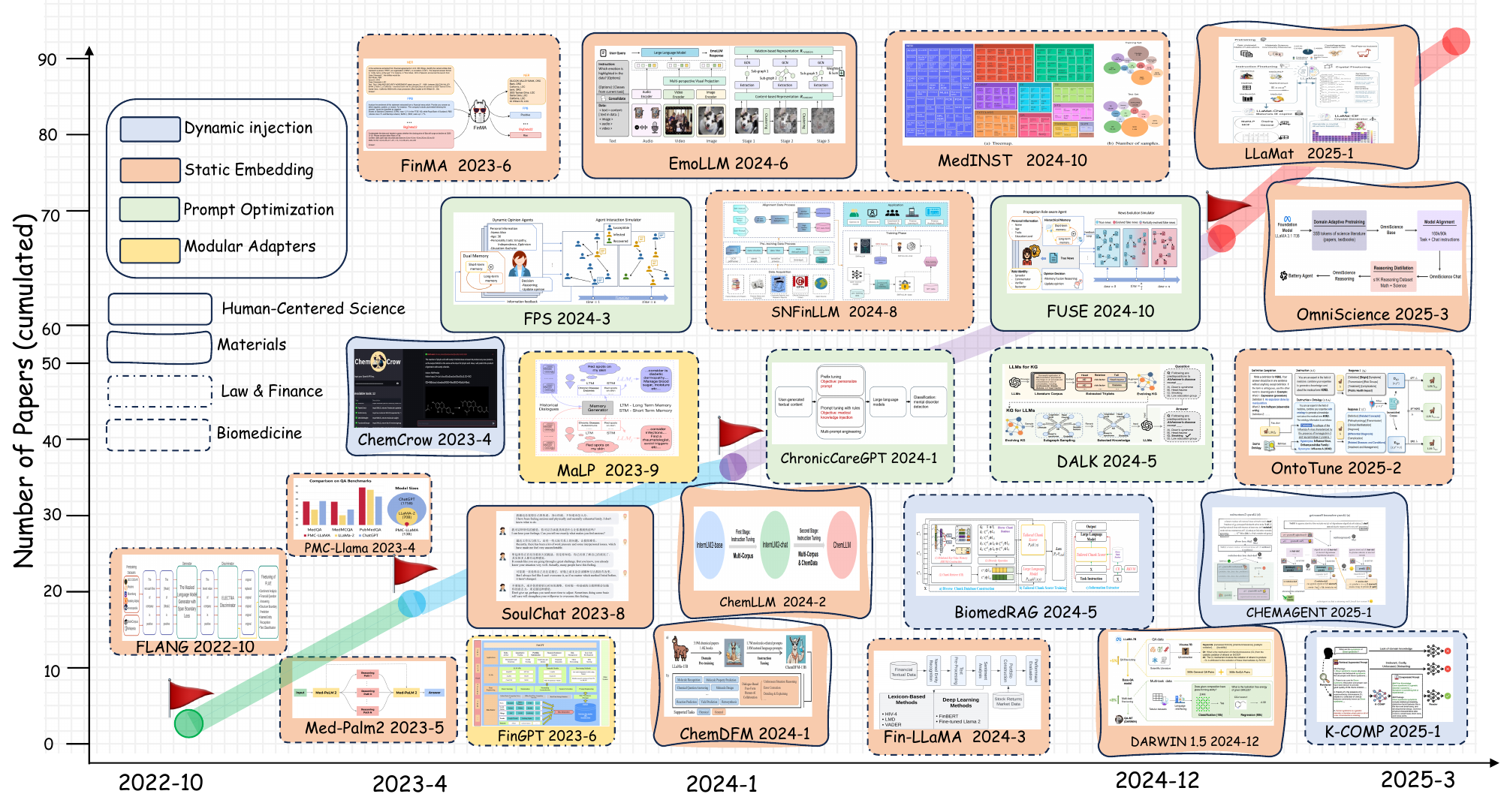}
    \caption{Illustration of Growth Trends in Domain-Specific Knowledge Injection into LLMs. 
The chart displays the cumulative number of papers published between October 2022 and December 2024.
Different colors and border styles represent various injection methods and domains. }
    \label{fig:intro}
\end{figure*}

LLMs have achieved extraordinary success across various tasks, showcasing remarkable capabilities in reasoning, knowledge representation, and decision-making~\cite{song2025maniplvm,xu2025socialmaze}. 
However, despite their impressive performance in general-purpose applications, many specialized domains, such as healthcare, chemistry, and legal analysis, demand the integration of domain-specific knowledge to achieve high accuracy and reliability. 
To address this challenge, researchers have explored methods to enhance LLMs through external or embedded domain expertise, a process often referred to as \textit{\textbf{knowledge injection}}, as shown in Figure~\ref{fig:intro}. This approach aims to bridge the gap between general-purpose language understanding and the stringent requirements of domain-specific tasks, enabling LLMs to perform effectively in highly specialized contexts.

Building on the foundational capabilities of general-purpose LLMs, knowledge injection techniques provide an effective means to address their limitations in handling specialized applications. Compared to the generalized approach of standard LLMs, knowledge injection offers two key advantages:  
1) incorporating precise, domain-specific knowledge to improve accuracy and reliability in specialized tasks, and  
2) allowing LLMs to dynamically adapt to new information or evolving knowledge bases, ensuring up-to-date expertise.  
These techniques bridge the gap between general-purpose understanding and domain-specific demands by leveraging both structured and unstructured knowledge sources. As a result, knowledge injection methods have been successfully applied in fields such as healthcare, chemistry, and legal analysis, significantly enhancing LLM performance. For example, biomedical LLMs ~\citep{cho2025kcompretrievalaugmentedmedicaldomain,bolton2024biomedlm27bparameterlanguage,yan2023biomedical} have demonstrated superior accuracy in tasks like medical diagnostics and regulatory compliance, while domain-specific models for material science~\citep{tang2025chemagentselfupdatinglibrarylarge,xie2024darwin15largelanguage,antunes2024crystalstructuregenerationautoregressive,zhang2024honeycombflexiblellmbasedagent} have achieved advances in material property prediction and discovery. These dedicated models underscore the transformative potential of integrating domain knowledge into LLMs.

Despite these advancements, early efforts in knowledge injection often treated domains independently, leading to a lack of standardization in methodologies and evaluation. 
As the volume of research continues to grow rapidly, with applications and studies proliferating across disciplines, the need for a comprehensive review becomes evident.
This review aims to summarize the state of knowledge injection techniques, provide a systematic blueprint for future research, and identify key challenges, such as balancing scalability with domain-specific accuracy and enabling efficient, real-time knowledge updates. 
We begin in Section~\ref{background} with background on domain-specific knowledge and its role in LLMs. Section~\ref{sec:paradigms} presents a unified framework of four knowledge injection paradigms: (1) Dynamic Knowledge Injection at inference time; (2) Static Knowledge Embedding during training or fine-tuning; (3) Modular Adapters for parameter-efficient integration; and (4) Prompt Optimization via carefully designed inputs. Section~\ref{application} examines these methods across domains such as materials science, chemistry, biology, and law. Section~\ref{resource} summarizes key datasets, tools, and comparative results. 
Section~\ref{challenges} outlines open challenges, including scalability, robustness, and domain transfer. Finally, Section~\ref{conclusion} concludes the paper and reflects on future directions.

\section{Background}
\label{background}

\subsection{Domain-Specific Knowledge}

Domain-specific knowledge refers to specialized information or expertise pertinent to a specific field or application, distinguishing it from general knowledge that spans across multiple domains. 
While general knowledge enables models to understand broad contexts, domain-specific knowledge is essential for addressing specialized tasks where precise, field-specific understanding is required~\cite{wang2025beyond,li2025flipping,zhang2025divide}.
For instance, in scientific text processing~\citep{bran2023chemcrow}, models must comprehend complex scientific terminologies, concepts, and methodologies to provide accurate and relevant answers. 
In e-commerce search~\citep{zhao2024snfinllm}, understanding domain-specific terms such as product categories, technical specifications, or colloquial shopping language is crucial for delivering relevant search results and recommendations.
In healthcare applications, LLMs must understand medical terminologies, diagnoses, treatment plans, and drug interactions. 
For example, biomedical question answering~\citep{singhal2025toward,pei2024biot5} and medical report summarization rely on integrating knowledge from medical literature like PubMed~\citep{dernoncourt2017pubmed}.
To address these needs, researchers have explored various methods for incorporating domain-specific knowledge into LLMs. 
In this paper, we aim to provide a survey of these various injection methods.

\subsection{Knowledge Representation and Encoding}
Knowledge can take different forms depending on structure and application needs. Knowledge graphs \citep{liao2025finetuningapproacht5using,zhang2024knowgpt} encode entities and their relationships in a structured graph, supporting reasoning and inference in tasks like question answering. In contrast, text-based sources like Wikipedia \citep{jeong2024adaptive} provide rich but unstructured information, useful for tasks requiring broad contextual understanding.
Knowledge can also be stored in vector space rather than in text or graph form. For example, soft prompt tuning~\citep{peng2025soft,singhal2023publisher} embeds useful information as vectors, which are appended to inputs to guide LLMs on specific tasks. Beyond external forms, knowledge may also emerge internally: chain-of-thought prompting~\citep{sanwal2025layeredchainofthoughtpromptingmultiagent,yao2024tree} introduces intermediate reasoning steps that help LLMs decompose complex problems and access internal knowledge more effectively—improving performance in tasks involving reasoning, multi-step computation, or decision-making.

\subsection{Knowledge Injection Survey}
Prior surveys on knowledge-enhanced language models vary in focus and scope. 
The most relevant works include the following: \citet{cadeddu2024comparative}, who focus on scientific article classification and offer practical insights but lack broader methodological generalization; \citet{wang2024knowledge}, who focus on knowledge editing and aim to update internal model knowledge with minimal side effects; and \citet{hu2023survey}, who adopt a model-centric perspective by classifying knowledge-enhanced models based on task type and knowledge source, though they primarily cover pre-LLM architectures such as BERT and ERNIE.
In contrast, our work presents a unified view of knowledge injection in LLMs, emphasizing capability enhancement through external knowledge integration across diverse tasks.

\section{Paradigms of Knowledge Injection}
\label{sec:paradigms}

\begin{figure*}[tb]
    \centering
    \includegraphics[width=1\linewidth]{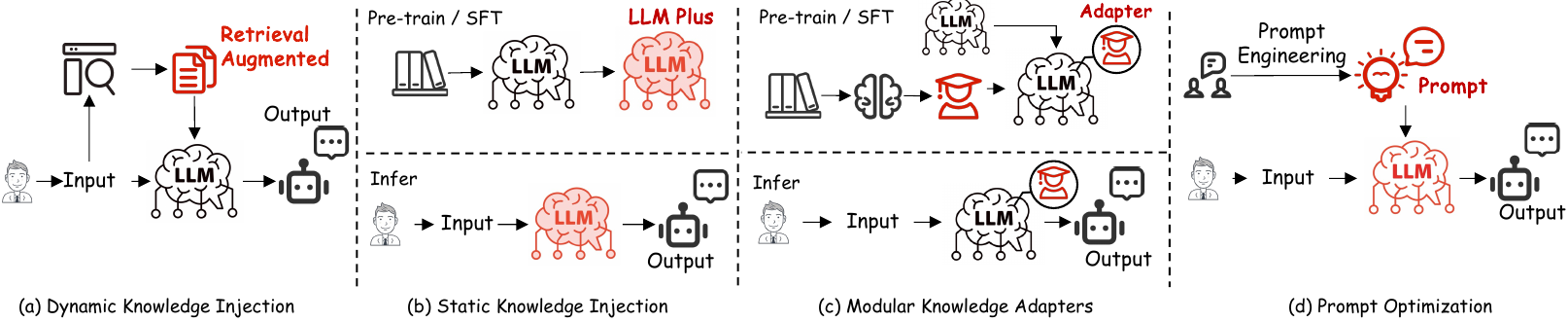}
    \caption{
Four knowledge injection paradigms for LLMs.
(a) Dynamic Knowledge Injection retrieves external knowledge during inference.
(b) Static Knowledge Injection embeds external knowledge into model parameters during fine-tuning.
(c) Modular Knowledge Adapters use plug-and-play modules to dynamically adapt to tasks.
(d) Prompt Optimization utilizes precise prompts to guide the LLM without altering its parameters.
    }
    \label{fig:model}
\end{figure*}

To systematically understand how domain knowledge is integrated into LLMs, we categorize existing approaches into four paradigms based on \textbf{when} the knowledge is incorporated and \textbf{how} it interacts with the model, as shown in Figure~\ref{fig:model}.
Specifically, \textit{Static Knowledge Injection} and \textit{Modular Knowledge Adapters} integrate knowledge prior to inference and involve parameter updates—through either full fine-tuning or adapter-based tuning.
In contrast, \textit{Dynamic Knowledge Injection} and \textit{Prompt Optimization} inject knowledge at inference time without altering model parameters: the former retrieves external information, while the latter leverages internal knowledge through designed prompts.

We utilize unified notations, as described in Table~\ref{form}, to systematically represent the processes.

\subsection{Dynamic Knowledge Injection}

\begin{table}[tb]
\centering
\small
\begin{tabular}{c|p{0.7\linewidth}}
\toprule
\textbf{Symbol} & \textbf{Description} \\
\midrule
\(\mathbf{x}\)  & Input to LLM \\ 
\(\mathbf{y}\)  & Output of LLM \\ 
\(M\) & Backbone LLM Function \\
\(\mathcal{K}\) & External domain knowledge base \\
\(\theta\)      & Parameters of LLM \\ 
\(\phi\)        & Additional parameters introduced \\ 
\(\mathcal{R}(\mathbf{x}, \mathcal{K})\) & Retrieval function fetches relevant elements of \(\mathcal{K}\) given the input \(\mathbf{x}\) \\
\(M(\mathbf{x}; \theta)\)& Represent LLM takes input \(\mathbf{x}\) and produces an output, parameterized by \(\theta\) \\
\(\Delta\theta\) & Offsets to the original LLM's parameters \\
\bottomrule
\end{tabular}

\caption{Summary of Symbols.}
\label{form}
\end{table}

We define dynamic knowledge injection as the process of first retrieving information from external knowledge bases or knowledge graphs and then combining it with the input for use in LLMs:
\begin{equation}
    \mathbf{y} = M(\mathbf{x}, \mathbf{\mathcal{R}(\mathbf{x}, \mathcal{K})}; \theta),
\end{equation}
where \(\mathbf{x}\) represents the original input, \(\mathcal{R}\) denotes the retrieval function, \(\mathcal{K}\) is the external knowledge base, and \(\theta\) are the model parameters, which remain unchanged. 
This paradigm offers several advantages, including ease of updating (hence the term "dynamic injection") and the ability to incorporate new knowledge without retraining the model.
However, it also presents challenges, such as dependency on the quality of the knowledge base \(\mathcal{K}\), the retrieval function \(\mathcal{R}\), and limitations imposed by the maximum input length of the LLM.
To improve retrieval quality, commonly used techniques include semantic matching based on sentence embeddings and efficient vector databases for fast similarity search.

\subsection{Static Knowledge Embedding}

Compared with dynamic knowledge retrieval, static knowledge embedding involves embedding knowledge into the model's parameters through full or partial fine-tuning, making it less flexible to changes.
Concretely, the model learns new parameters \(\Delta\theta\) that encode domain knowledge from \(\mathcal{K}\):
\[
\Delta\theta = \textstyle \arg\min_{\theta} \sum_{(\mathbf{x_s}, \mathbf{y_s}) \in \mathcal{K}} \mathcal{L}\bigl(M(\mathbf{x_s}; \theta), \mathbf{y_s}\bigr),
\]
where \(\mathcal{K}\) is the domain-specific knowledge base containing training samples \(\mathbf{x_s}\) and \(\mathbf{y_s}\), and \(\mathcal{L}\) is a training loss function. After optimization, the updated parameters \(\Delta\theta\) are obtained. 

At inference time, no further retrieval or external knowledge calls are required: $\mathbf{y} = M\bigl(\mathbf{x}; \Delta\theta\bigr)$.
This paradigm enables fast inference by removing the need for additional retrieval steps and often delivers stronger performance. However, it also presents challenges, such as high update costs since fine-tuning is required when domain knowledge changes, and scalability concerns because embedding large or frequently changing knowledge bases demands significant computational resources.

\subsection{Modular Knowledge Adapters}  
To address the costly updates associated with static knowledge embedding, another paradigm, known as modular knowledge adapters, introduces \textit{small}, trainable modules that can be inserted into or operate alongside the base model to store domain-specific knowledge while saving computational resources. 
In this approach, the original parameters \(\theta\) of the LLM typically remain frozen, preserving the model’s general-purpose capabilities.
Given a knowledge dataset \(\mathcal{K}\), the adapter parameters \(\phi\) are trained by minimizing the following objective:
\[
\phi = \textstyle \arg\min_{\phi} \sum_{(\mathbf{x_s}, \mathbf{y_s}) \in \mathcal{K}} \mathcal{L}\bigl(M(\mathbf{x_s}; \theta, \phi), \mathbf{y_s}\bigr),
\]
where \(M(\mathbf{x_s}; \theta, \phi)\) represents the base model’s generation function enhanced with the new adapter parameters. 
At inference time, the enhanced model generates outputs as: $\mathbf{y} = M\bigl(\mathbf{x}; \theta, \phi\bigr)$.
This paradigm offers a parameter-efficient method to adapt LLMs to specific domains without modifying the original model weights. 
By freezing the base model’s parameters, the approach seeks to preserve previously acquired knowledge while enabling the seamless incorporation of new domain-specific information.
However, this approach also introduces challenges, such as the need to design new architectural components and determine appropriate hyperparameters, including the size and number of adapters. These additional elements can increase the overall complexity of the model and its training process.

\subsection{Prompt Optimization}  
Prompt optimization refers to the practice of guiding LLMs to perform domain-specific tasks by crafting effective textual prompts.
Unlike retrieval-based methods, it relies entirely on the model’s internal knowledge and does not require access to external knowledge bases or fine-tuning. 
The process can be formalized as:
\[
\mathbf{p}^* =\textstyle  \arg\min_{\mathbf{p}} \mathcal{L}\big(M([\mathbf{p}, \mathbf{x}]; \theta), \mathbf{y}^*\big),
\]
where \(\mathbf{p}\) is a prompt containing domain-relevant cues, \(\mathbf{x}\) is the task input, and \(\theta\) are the fixed parameters of the LLM.

This paradigm offers practical advantages such as lightweight deployment, no training overhead, and adaptability across domains. However, it also faces key challenges: designing prompts that elicit accurate responses can be non-trivial, and long prompts may reduce the available input space due to context length limitations.
Prompt-based approaches can be broadly categorized into manual prompting, prompt tuning, and prefix tuning. 
These differ in how prompts are constructed or optimized—ranging from discrete, static prompts to learnable embeddings—and have been widely adopted for low-resource domain adaptation.

\subsection{Comparison of the Four Paradigms}

\begin{table}[htbp]
\centering
\scriptsize
\renewcommand{\arraystretch}{1} 
\begin{tabular}{@{}m{1cm}m{1.8cm}m{1.5cm}m{2.2cm}@{}}

\toprule
\textbf{Paradigm} & \textbf{Training Cost} & \textbf{Inference Speed} & \textbf{Limitations} \\ \midrule
\begin{tabular}[c]{@{}c@{}}Dynamic \\ Injection\end{tabular}  &   \begin{tabular}[c]{@{}c@{}} None, but requires \\ retrieval module\end{tabular}&  \begin{tabular}[c]{@{}c@{}} Slower due to  \\ retrieval latency \end{tabular}& \begin{tabular}[c]{@{}c@{}} Relies heavily on\\retrieval quality\end{tabular} \\ \midrule
\begin{tabular}[c]{@{}l@{}}Static \\ Embedding\end{tabular}   & \begin{tabular}[c]{@{}cc@{}} High \\ (requires pretraining \\ or fine-tuning)\end{tabular} & No extra cost & \begin{tabular}[c]{@{}cc@{}} Fixed knowledge; \\ risks catastrophic \\ forgetting \end{tabular}  \\ \midrule
\begin{tabular}[c]{@{}c@{}}Modular \\ Adapters\end{tabular}   & \begin{tabular}[c]{@{}cc@{}} Low \\ (train small subset \\ of parameters)\end{tabular} & \begin{tabular}[c]{@{}c@{}} Almost  \\unaffected\end{tabular} & \begin{tabular}[c]{@{}c@{}} Sensitive to training \\ data quality\end{tabular} \\ \midrule
\begin{tabular}[c]{@{}l@{}}Prompt \\ Optimization\end{tabular} & None & \begin{tabular}[c]{@{}c@{}} Almost  \\unaffected\end{tabular} & \begin{tabular}[c]{@{}cc@{}} Labor-intensive;\\ limited to pre-existing\\ knowledge\end{tabular}  \\
\bottomrule
\end{tabular}
\caption{Guidance on choosing injection methods by training cost, inference speed, and constraints.}
\label{tab:knowledge_injection}
\end{table}

Dynamic knowledge injection introduces external knowledge at runtime, offering flexibility and adaptability without added training cost. However, it depends on efficient retrieval, and inference speed can suffer if retrieval performance is poor.
Static knowledge embedding integrates domain expertise during pretraining or fine-tuning, requiring large-scale data and significant computational resources. It adds no inference cost but struggles to adapt to new information and is prone to catastrophic forgetting.
Modular adapters offer a middle ground by enhancing domain capabilities through plug-and-play modules that require minimal training data. Only a small number of parameters are trained, reducing cost and preserving inference speed, though performance heavily depends on data quality.
Prompt optimization avoids retraining by using well-crafted inputs. It maintains fast inference but relies on significant manual effort and is limited to activating existing knowledge rather than incorporating new information.
We summarize these comparisons in Table~\ref{tab:knowledge_injection} as a practical guide to help determine the suitable method based on specific requirements and scenarios.

\section{Applications}
\label{application}

\subsection{Finance}  

In the financial domain, LLM generally follow two main development paths: fine-tuning general-purpose models on financial tasks or training models from scratch using domain-specific corpora.

For fine-tuning, PIXIU\citep{xie2023pixiu} adapts LLaMA using 136K financial instruction samples, equipping the model to handle diverse finance-related scenarios. Instruct-FinGPT\citep{zhang2023instruct} focuses on sentiment classification by fine-tuning on 10K samples from two financial sentiment datasets. FinGPT~\citep{yang2023fingpt} proposes an end-to-end framework for developing FinLLMs, efficiently fine-tuning LLaMA and ChatGLM with 50K samples via LoRA, significantly reducing computational costs.
In contrast, scratch-trained FinLLMs aim for deep domain alignment from the ground up. BloombergGPT \citep{wu2023bloomberggpt} uses 5B Bloomberg-specific tokens (0.7\% of its total corpus) to specialize in financial applications. XuanYuan 2.0 \citep{zhang2023xuanyuan} is the largest Chinese financial chatbot, trained on 366B tokens and fine-tuned on 13B. Fin-T5 \citep{lu2023bbt} leverages a 300GB Chinese financial corpus using the T5 architecture, while SNFinLLM \citep{zhao2024snfinllm} enhances inference through real-time financial data injection.

In summary, this field showcases a rich diversity in training strategies, from lightweight tuning to comprehensive, end-to-end development.

\begin{table*}[h!]
\scriptsize
\resizebox{\textwidth}{!}{%
\begin{tabular}{l|c|c|c|c}
\toprule
\textbf{Domain} & \textbf{Model} & \textbf{Paradigms} & \textbf{Knowledge Source} & \textbf{Link}  \\ 

 \hline
\multirow{8}{*}{Finance}
&FLANG    ~\citep{shah2022fluemeetsflangbenchmarks}&Static Knowledge Embedding   &\begin{tabular}[c]{@{}c@{}}Financial PhraseBank,FiQA 2018 Task-1,\\News Headline Classification, Named Entity Recognition,\\Structure Boundary Detection,Question Answering \end{tabular} &\href{https://huggingface.co/SALT-NLP/FLANG-BERT}{\textcolor{teal}{Link}} \\ \cline{2-5}

& BloomBergGPT~\citep{wu2023bloomberggpt} & \begin{tabular}[c]{@{}c@{}}Static Knowledge Embedding\end{tabular} & \begin{tabular}[c]{@{}c@{}}Finance dataset (web, news, filings, press, Bloomberg), \\
Public dataset (the Pile, C4, Wikipedia)\end{tabular} & \textbackslash{} \\ 
\cline{2-5}


&FinMA ~\citep{xie2023pixiu}&Static Knowledge Embedding   &\begin{tabular}[c]{@{}c@{}}FPB,FiQA-SA,Headline,NER,FinQA,\\ConvFinQA,BigData22,ACL18,CIKM18    \end{tabular}
  &\href{https://github.com/The-FinAI/PIXIU}{\textcolor{teal}{Link}} \\ \cline{2-5}

&FinGPT~\citep{zhang2023instruct} & Modular Knowledge Adapters & \begin{tabular}[c]{@{}c@{}}Financial news, Company filings and announcements, \\
 Social media discussions, Trends\end{tabular}
 &\href{https://github.com/AI4Finance-Foundation/FinGPT}{\textcolor{teal}{Link}}  \\ 
\cline{2-5}

 &Fin-LLaMA  ~\citep{konstantinidis2024finllamafinancialsentimentclassification}&Static Knowledge Embedding   &fin-llama-dataset   &\href{https://github.com/Bavest/fin-llama}{\textcolor{teal}{Link}} \\ \cline{2-5}

&SNFinLLM~\citep{zhao2024snfinllm}  & \begin{tabular}[c]{@{}c@{}}Static Knowledge Embedding\end{tabular} & \begin{tabular}[c]{@{}c@{}}FinEval, FinanceIQ,qEQA,FinC,KQA,MRC,cMRC\end{tabular}
 &\textbackslash{} \\ 
\cline{2-5}

&Fino1 ~\citep{qian2025fino1transferabilityreasoningenhanced}&Static Knowledge Embedding   &
\begin{tabular}[c]{@{}c@{}}FinQA, TATQA , DocMath (Simpshort and Compshort),\\
 DocFinQA , and BizBench-QA\end{tabular}
&\href{https://github.com/The-FinAI/Fino1}{\textcolor{teal}{Link}} \\ \cline{1-5}



\hline
\multirow{10}{*}{Biomedicine}   

&PMC-LLaMA~\citep{wu2023pmc}& Static Knowledge Embedding &PMC-OA, MedC-I, PubMedQA, MedMCQA, USMLE& \href{https://github.com/chaoyi-wu/PMC-LLaMA}{\textcolor{teal}{Link}}  \\ \cline{2-5}

&Med-PaLM 2~\citep{singhal2023towards}&Static Knowledge Embedding&MultiMed&\href{https://github.com/SHARANR26/Med-Palm2}{\textcolor{teal}{Link}}  \\ \cline{2-5}


&DALK ~\citep{li2024dalk}&\begin{tabular}[c]{@{}c@{}}Dynamic Knowledge Injection \\ Prompt Optimization\end{tabular}&MedQA, MedMCQA, MMLU, QA4MRE&\href{https://github.com/David-Li0406/DALK}{\textcolor{teal}{Link}} \\ \cline{2-5}

&ChronicCareGPT~\citep{liu2024few} &Prompt Optimization& eRisk &\href{https://github.com/WangRongsheng/CareGPT}{\textcolor{teal}{Link}} \\ \cline{2-5}

&SA-MDKIF~\citep{xu2024sa}&Modular Knowledge Adapters&MedQuA,emrQA, PubMedQA, MedQA& \textbackslash{}\\ \cline{2-5}

&MaLP~\citep{zhang2024llm}&Modular Knowledge Adapters&HealthCareMagic-100k, iCliniq &\href{https://github.com/MatthewKKai/MaLP}{\textcolor{teal}{Link}} \\ \cline{2-5}
&BioMedLM ~\citep{bolton2024biomedlm27bparameterlanguage}&Static Knowledge Embedding   &PubMed,MedMCQA,MedQA,MMLU,BioASQ   &\href{https://github.com/stanford-crfm/BioMedLM}{\textcolor{teal}{Link}} \\ \cline{2-5}

&BiomedRAG~\citep{li2024biomedragretrievalaugmentedlarge}&Dynamic Knowledge Injection    &CHEMPROT,DDI,ade-corpus-v2,MTsample,ADInt,UMLS   &\href{https://github.com/ToneLi/BIoMedRAG}{\textcolor{teal}{Link}} \\ \cline{2-5}

&MedINST~\citep{han2024medinstmetadatasetbiomedical}&Static Knowledge Embedding   &MedINST   &\href{https://github.com/MedMNIST/MedMNIST}{\textcolor{teal}{Link}} \\ \cline{2-5}

&K-COMP~\citep{cho2025kcompretrievalaugmentedmedicaldomain}&Dynamic Knowledge Injection   &MedCorp corpus   &\textbackslash{}  \\ \cline{2-5}

&OntoTune~\citep{liu2025ontotuneontologydrivenselftrainingaligning}&Static Knowledge Embedding   &SemEval2018 Task 9  dataset&\href{https://github.com/zjukg/OntoTune}{\textcolor{teal}{Link}} \\ \cline{1-5}

\hline

\multirow{8}{*}{Materials}   

 & ChemCrow~\citep{bran2023chemcrow} & Dynamic Knowledge Injection & 18 expert-designed tools & \href{https://github.com/ur-whitelab/chemcrow-public}{\textcolor{teal}{Link}} \\
  \cline{2-5}

 &  ChemDFM ~\citep{zhao2024chemdfmlargelanguagefoundation} &Static Knowledge Embedding & \begin{tabular}[c]{@{}c@{}}SciQ,PIQA,PubChem,ARC,USPTO\end{tabular} &\href{https://github.com/OpenDFM/ChemDFM}{\textcolor{teal}{Link}} \\
 \cline{2-5}

 & ChemLLM~\citep{zhang2024chemllmchemicallargelanguage} &\begin{tabular}[c]{@{}c@{}}Static Knowledge Embedding\end{tabular} & \begin{tabular}[c]{@{}c@{}}ChemData,ChemBench\end{tabular} & \href{https://github.com/keyhsw/ChemLLM}{\textcolor{teal}{Link}} \\
 \cline{2-5}

 & CrystaLLM ~\citep{antunes2024crystalstructuregenerationautoregressive} &\begin{tabular}[c]{@{}c@{}}Static Knowledge Embedding\end{tabular} & \begin{tabular}[c]{@{}c@{}}Materials Project, OQMD, NOMAD\end{tabular} & \href{https://github.com/lantunes/CrystaLLM}{\textcolor{teal}{Link}} \\
 \cline{2-5}
 & ScholarChemQA ~\citep{chen2024scholarchemqa} &\begin{tabular}[c]{@{}c@{}}Static Knowledge Embedding\end{tabular} & AG News,Yahoo Answers ,Yelp-5,Amazon-5 & \href{https://github.com/iriscxy/chemmatch}{\textcolor{teal}{Link}} \\
 \cline{2-5}

& DARWIN 1.5~\citep{xie2024darwin15largelanguage} & Static Knowledge Embedding & FAIR datasets &  \href{https://github.com/MasterAI-EAM/Darwin}{\textcolor{teal}{Link}} \\ 
\cline{2-5}

  &  ChemAgent~\citep{tang2025chemagent} &\begin{tabular}[c]{@{}c@{}}Dynamic Knowledge Injection \\ Prompt Optimization\end{tabular} & \begin{tabular}[c]{@{}c@{}}
  college chemistry textbooks:Quantum chemistry (quan), hemistry kinetics (matter)\\ Quantum mechanics (chemmc), Physical chemistry (atkins)\end{tabular} & \href{https://github.com/gersteinlab/chemagent}{\textcolor{teal}{Link}} \\
 \cline{2-5}

&  LLaMat
~\citep{mishra2025foundationallargelanguagemodels} &Static Knowledge Embedding & MatBookQA,MaScQA,MatSciInstruct &  \textbackslash{} \\
 \cline{2-5}

 &  OmniScience~\citep{prabhakar2025omnisciencedomainspecializedllmscientific} &Static Knowledge Embedding &daring-anteater dataset, s1K dataset & \textbackslash{} \\
 \cline{1-5}
 

\hline

\multirow{5}{*}{Mental Health}  
 &  MeChat~\citep{qiu2023smile} & Dynamic Knowledge Injection  & SMILECHAT, PsyQA & \href{https://huggingface.co/qiuhuachuan/MeChat}{\textcolor{teal}{Link}}  \\  \cline{2-5}

  &  MindChat~\citep{MindChat} & Static Knowledge Embedding & Multi-turn psychological dialogue data & \href{https://github.com/X-D-Lab/MindChat}{\textcolor{teal}{Link}}  \\  \cline{2-5}

 &  SoulChat~\citep{chen2023soulchat} & Static Knowledge Embedding & Long-text counseling sessions & \href{https://github.com/scutcyr/SoulChat}{\textcolor{teal}{Link}}  \\ \cline{2-5}


  & EmoLLM~\citep{yang2024emollm} & \begin{tabular}[c]{@{}c@{}}Static Knowledge Embedding \\ Modular Knowledge Adapters\end{tabular} &  CPsyCounD & \href{https://github.com/SmartFlowAI/EmoLLM}{\textcolor{teal}{Link}}  \\

  \hline
  \multirow{5}{*}{Education}  
  &  EduChat~\citep{dan2023educhat} & Static Knowledge Embedding & \begin{tabular}[c]{@{}c@{}}Textbooks Data, Open QA Data, \\
 Emotional Support Data, Socratic Teaching Data\end{tabular} & \href{https://github.com/ECNU-ICALK/EduChat}{\textcolor{teal}{Link}}  \\  \cline{2-5}

&  QiaoBan~\citep{qiaoban2023} & Prompt Optimization & Children's emotional education dialogue data & \href{https://github.com/HIT-SCIR-SC/QiaoBan?tab=readme-ov-file}{\textcolor{teal}{Link}}   \\ \cline{2-5}
      
  &  HiTA~\citep{liu2024hita} & Dynamic Knowledge Injection & Educator curated database & \textbackslash{}  \\  \cline{2-5}

& SocraticLM~\citep{liusocraticlm} & Modular Knowledge Adapters & SocraTeach dataset& \textbackslash{}  \\ \cline{2-5}

& CyberQ~\citep{agrawal2024cyberq} & \begin{tabular}[c]{@{}c@{}}Static Knowledge Embedding \\ Dynamic Knowledge Injection\end{tabular} &  AISecKG, Q\&A  & \textbackslash{}  \\ 
\hline

    \multirow{3}{*}{Social Science} 
  & SocialLLM~\citep{jiang2023social}  &  
\begin{tabular}[c]{@{}c@{}}Static Knowledge Embedding \\
 Prompt Optimization\end{tabular}& \begin{tabular}[c]{@{}c@{}}Covid-Political, Election2020, COVID-Morality, \\ Ukr-Rus-Suspended, Ukr-Rus-Hate,  \\ Immigration-Hate-08, Immigration-Hate-05\end{tabular} & \textbackslash{} \\ \cline{2-5}
 
  & FPS~\citep{liu2024skepticism}  &  
Prompt Optimization & Fake News Dataset, Big Five Personality Traits & \href{https://github.com/LiuYuHan31/FPS}{\textcolor{teal}{Link}} \\ \cline{2-5}

  & FUSE~\citep{liu2024tiny}  &  
Prompt Optimization & True News Dataset, Big Five Personality Traits & \textbackslash{} \\
\toprule
\end{tabular}}
\caption{Summary of the domain-specific knowledge injection studies. 
We categorize current work according to their research domain and knowledge injection method.}
\label{maintable}
\end{table*}

\subsection{Biomedicine}

The biomedicine domain benefits from a wealth of specialized corpora, such as PubMed~\citep{dernoncourt2017pubmed} and MedQA~\citep{jin2021disease}, enabling the development of LLMs specifically trained on biomedical texts. These models often follow the static knowledge embedding approach, leveraging the domain-specific richness of biomedical data.
For instance, PMC-LLaMA~\citep{wu2023pmc} extends the LLaMA 7B model through further pretraining on 4.9 million PubMed Central articles curated from the S2ORC dataset~\citep{lo2020s2orc}, completing five epochs to embed biomedical knowledge effectively. Similarly, Med-PaLM 2~\citep{singhal2023towards} builds on PaLM 2 via instruction fine-tuning. This fine-tuning incorporates a diverse mix of medical question-answering datasets, including MedQA, MedMCQA~\citep{pal2022medmcqa}, and HealthSearchQA~\citep{singhal2023publisher}.

Beyond foundational models, integrating external tools and knowledge can further enhance performance. GeneGPT\citep{jin2024genegpt} leverages a code-pretrained LLM to address GeneTuring tests by calling NCBI Web APIs, combining in-context learning with an augmented decoding algorithm capable of identifying and executing API requests. Med-PaLM\citep{singhal2023publisher} extends the capabilities of Flan-PaLM~\citep{chung2024scaling} through the use of vector prompts—dense representations designed to store and retrieve medical domain knowledge during inference.

Overall, biomedical LLMs lead in combining static pretraining, instruction tuning, and tool integration, reflecting a shift toward hybrid reasoning in specialized AI.

\subsection{Materials}
In contrast to the biomedical domain, the field of materials science and chemistry has largely focused on static knowledge embedding. Many models rely on domain-specific corpora to fine-tune general models for improved task performance.
Darwin 1.5 \citep{xie2024darwin15largelanguage} adopts a two-stage training strategy using natural language inputs to enhance performance in materials discovery. ScholarChemQA \citep{chen2024scholarchemqa} constructs a chemistry QA dataset to fine-tune BERT and LLaMA, improving chemical reasoning.
Recently, some efforts have begun to explore dynamic knowledge integration. ChemCrow \citep{bran2023chemcrow} augments LLMs with chemistry tools for applications like synthesis and drug discovery. ChemAgent \citep{tang2025chemagent} shows that well-designed planning prompts can guide models through complex execution tasks by leveraging internal reasoning.

 While still in early stages, the field is transitioning from static embedding toward interactive and tool-augmented reasoning, indicating strong potential for future developments

\subsection{Human-Centered Science}  

Human-centered science focuses on understanding and assisting human behaviors, needs, and decisions. This interdisciplinary domain includes mental health, education, social behavior prediction, and legal reasoning—each benefiting from personalized and context-aware LLMs.

In \textit{\textbf{mental health}}, datasets like PsyQA~\citep{sun2021psyqa} provide a foundation for training models in psychological counseling scenarios.
SoulChat~\citep{chen2023soulchat}, a model fine-tuned on over 100,000 long-text counseling sessions using static knowledge embedding, is designed for empathic conversations. 
In contrast, MeChat~\citep{qiu2023smile} employs dynamic knowledge injection to adapt to real-time inputs, enhancing its emotional support capabilities. 
These advancements demonstrate the potential of human-centered science in addressing complex, real-world challenges through personalized and context-aware solutions.  

In the \textit{\textbf{education domain}}, LLMs have shown immense potential in addressing challenges such as personalized learning, curriculum alignment, and interactive teaching. 
Personalized learning, for example, requires models to adapt to individual needs, providing tailored feedback and emotional support. 
EduChat~\citep{dan2023educhat} applies psychological and pedagogical theories via static knowledge embedding to support tasks like Q\&A, writing feedback, and emotional guidance. Similarly, QiaoBan~\citep{qiaoban2023} uses prompt optimization to tailor model behavior to children's psychological and emotional needs.
Domain-specific education and interactive teaching have also seen advancements. 
CyberQ~\citep{agrawal2024cyberq} blends static knowledge embedding and dynamic knowledge injection via AISecKG~\citep{agrawal2023aiseckg}, generating Q\&A based on cybersecurity best practices.
Interactive teaching, on the other hand, benefits from models like SocraticLM~\citep{liusocraticlm}, which employs adapters fine-tuned on the SocraTeach dataset to engage students in critical thinking and problem-solving. 

For \textit{\textbf{social sciences}}, models like SocialLLM~\citep{jiang2023social} combine static knowledge embedding and dynamic knowledge injection to analyze human behavior in social networks. 
Models like SSF~\citep{wang2024decoding} FPS~\citep{liu2024skepticism} and FUSE~\citep{liu2024tiny} use prompt optimization to simulate the spread and evolution of fake news in social networks, helping understand misinformation's impact. In addition, \cite{liu2025truth} adopts a multi-agent approach to synthesize data, which is then used as knowledge to enhance language models for detecting fake news.
A summary of the mainstream models and their information is provided in Table~\ref{maintable}.
More models across various domains can be found at: \href{https://github.com/abilliyb/Knowledge_Injection_Survey_Papers}{\textcolor{blue}{official-repo}}.

\section{Tools, Resources, and Analysis}
\label{resource}

\subsection{Knowledge Injection Framework}
In this section, we provide a detailed introduction to four open-source frameworks categorized under different knowledge injection methods to facilitate understanding and application:
KnowGPT~\citep{zhang2024knowgpt} for Dynamic Knowledge Injection, StructTuning~\citep{liu2024structure} for Static Knowledge Embedding, 
K-Adapter~\citep{wang2021k} for Modular Knowledge Adapters, and SelfLift~\citep{cheng2024lift} for Prompt Optimization.

KnowGPT \textit{dynamically} combines knowledge graphs with prompt optimization by leveraging reinforcement learning to extract highly relevant subgraphs from the knowledge graph. 
These subgraphs are represented as triples and transformed into natural language prompts that language models can interpret and utilize via diverse prompt templates. 
The KnowGPT framework significantly reduces the API call costs of LLMs while enhancing their performance in domain-specific tasks.


StructTuning uses a structure-aware approach to \textit{embed} domain knowledge into pre-trained models with a two-stage strategy: 
Structure-Aware Continual Pre-Training encodes knowledge into the model's parameters, and Structure-Aware Supervised Fine-Tuning refines understanding through structured QA tasks.
This framework demonstrates significant performance improvements in knowledge-driven tasks such as relation classification and question answering, achieving a balance between generality and efficiency.

K-Adapter stores knowledge within \textit{adapter} modules. 
Its core method involves freezing the original model parameters and assigning an independent, task-specific adapter for each type of knowledge.
These adapters are inserted as independent modules into the intermediate layers of the model to generate enhanced representations of specific knowledge. 
This design effectively mitigates the issue of catastrophic forgetting, preventing newly injected knowledge from overwriting the model's pre-existing knowledge.

Finally, SelfLift iteratively employs a retrieval-augmented generator to create an unbounded memory pool and uses a memory selector to choose one output as memory for the subsequent generation round. 
This is a good demonstration of prompt optimization, where the model's outputs are dynamically refined and reused to enhance its overall performance and coherence in subsequent tasks.

\subsection{Knowledge Source}
We summarize commonly used knowledge sources for domain-specific LLMs in Table~\ref{maintable}, referring to datasets that provide the external knowledge used in various injection methods—including training corpora for static embedding or adapter tuning, and retrieval or prompt design resources for dynamic knowledge injection. 
Biomedicine includes numerous high-quality datasets, such as PubMed, PubMedQA~\citep{jin2019pubmedqa}, and BioASQ~\cite{tsatsaronis2012bioasq}, which support tasks such as question answering and clinical summarization. 
In contrast, materials and chemistry have more limited resources, and datasets like USPTO and Enzymes focus on chemical reactions.
Miscellaneous datasets are scattered across other domains, such as PsyQA and SmileChat in mental health, SocraTeach, and Children’s emotional education dialogue data dataset in education.
This diversity underscores the effort to tailor LLMs to specialized fields while emphasizing the need for broader curation of benchmarks in underrepresented domains.

\subsection{Performance Comparison of 4 Paradigms}
\begin{table}[htbp]
\centering
\small
\resizebox{0.48\textwidth}{!}{%
\begin{tabular}{@{}lcccc@{}}
\toprule
\textbf{Model}          & \textbf{Category} & \textbf{MedQA} & \textbf{PubMedQA} & \textbf{MedMCQA} \\ \midrule
GPT-4 (Medprompt)       & Prompt Optimization & 90.2           & 82.0              & 79.1             \\
GPT-4                  & General              & 90.2           & 80.4              & 73.7             \\
Med-PaLM 2             & Static Knowledge     & 85.4           & 81.8              & 72.3             \\
Flan-PaLM (3-shot)     & Dynamic Knowledge    & 67.6           & 79.0              & 57.6             \\
PMC-LLaMA              & Static Knowledge     & 56.3           & 77.9              & 56.0             \\
BioMedLM               & Static Knowledge             & 50.3           & 74.4              & --               \\
LLaMA (MedAdapter)     & Knowledge Adapters   & 37.4         & 63.6              & 32.0             \\

\bottomrule
\end{tabular}%
}
\caption{Model performance across four knowledge paradigms on medical benchmarks.}
\label{compare}
\end{table}

To compare knowledge injection paradigms in a practical setting, we focus on the biomedical domain due to its popularity and the availability of benchmarks such as MedQA, PubMedQA, and MedMCQA, as shown in Table~\ref{compare}.
Although the models differ in architecture, we align backbones when possible. For example, both PMC-LLaMA and MedAdapter use LLaMA-13B.
SOTA models like GPT-4 are closed-source, making prompt optimization the only feasible adaptation strategy. 
Despite no domain-specific training, GPT-4 with Medprompt achieves strong performance, showing the effectiveness of prompt methods for closed models.
Among open models, MedAdapter underperforms compared to PMC-LLaMA, suggesting that full fine-tuning may outperform adapter-based methods for some tasks.
Performance differences across paradigms also highlight the importance of pretraining corpus and task alignment, particularly in static injection approaches.
Furthermore, in Appendix~\ref{4paradigm}, we also compare knowledge injection paradigms in the finance domain and obtain similar conclusions to those in the medical domain.

\section{Challenges and Opportunities}
\label{challenges}
\textbf{Integrated Knowledge Consistency.}
Knowledge injection allows LLMs to incorporate and integrate different domain-specific knowledge. 
However, retrieved knowledge may conflict with the model’s pre-trained representations or other retrieved facts, leading to inconsistencies in outputs~\citep{xu2024knowledge}. 
For example, in healthcare or legal analysis, conflicting treatment protocols or contradictory legal precedents could arise~\cite{dayton2012standards}, resulting in unreliable decisions and undermining the system’s trustworthiness.
To address this, future research must focus on detecting inconsistencies, resolving conflicts, and maintaining consistency in integrated knowledge.
Conflicts can be addressed by prioritizing reliable sources, applying domain-specific rules, or using ensemble techniques to balance multiple perspectives. 
Alignment and validation modules help ensure retrieved knowledge fits the model’s reasoning.

\textbf{Cross-Domain Knowledge Transfer.}
Cross-domain knowledge transfer involves equipping LLMs with the ability to generalize knowledge across diverse and distinct fields~\cite{li2025zero}. 
While this significantly expands their applicability, it also introduces challenges due to the complexity and diversity of domain-specific terminologies, ontologies, and reasoning patterns~\cite{montero2004semantic}.
For example, transferring knowledge from chemistry to healthcare might require reconciling differing data structures and reasoning~frameworks~\citep{schroeder2018relating}.
Overcoming these challenges requires advancements in modular knowledge representation and transfer learning techniques. 
Future efforts could explore hybrid approaches that blend static embeddings with dynamic retrieval, enabling LLMs to adapt knowledge flexibly across domains without compromising depth. 
Additionally, standardized cross-domain benchmarks can enable consistent evaluation and drive innovation in knowledge transfer methods.
We provide more discussions in Appendix~\ref{challenges}.

\section{Conclusion}
\label{conclusion}
LLMs enhanced by domain-specific knowledge have shown remarkable potential and garnered increasing research interest. 
This survey systematically reviews LLM knowledge injection systems, exploring knowledge representation methods, integration strategies, and mechanisms for preserving model generality. 
We also summarize applications across biomedicine, chemistry, and computational social science domains.
By highlighting standard datasets, benchmarks, challenges, and future opportunities, we aim to provide a valuable resource that inspires exploration of knowledge-enhanced LLMs for domain-specific challenges.

\section*{Limitation}
Despite providing a comprehensive review of current methods and applications for domain-specific knowledge injection in LLMs, this survey has certain limitations.
First, while we strive to cover several key domains such as finance, biomedicine, and materials science, some less-studied or emerging areas (for example, low-resource languages, cross-cultural education, and niche disciplines) receive relatively limited attention. 
Second, our focus is primarily on summarizing methodological principles and representative models from existing literature.
Due to substantial variation in model architectures, application domains, training data, and evaluation protocols, we were only able to conduct targeted comparisons under controlled conditions within the biomedical domain, using commonly adopted datasets.
A more systematic and broad-based empirical evaluation across methods remains an important direction for future work. 
Nevertheless, we hope this survey serves as a useful reference and provides a clear roadmap for ongoing research in knowledge-enhanced LLMs.

\section*{Acknowledgement}

We would like to thank the anonymous reviewers for their constructive comments. 
The work was supported by Mohamed bin Zayed University of Artificial Intelligence (MBZUAI) through grant award 8481000078.

\bibliography{custom}

\appendix

\section{Performance Comparison of 4 Paradigms}
\label{4paradigm}


\begin{table}[htbp]
\centering
\small
\resizebox{0.48\textwidth}{!}{%
\begin{tabular}{@{}lcccc@{}}
\toprule
\textbf{Model}          & \textbf{Category} & \textbf{FPB} & \textbf{FiQA-SA} & \textbf{TFNS} \\ \midrule
GPT-4       & General  &83.3           &63.0              &80.8             \\

GPT-4(finetune)       & Static Knowledge  &87.8           &88.7              &88.3             \\
FinGPT        &Knowledge Adapters   &88.2           &87.4              &90.3             \\

FinBERT       &Static Knowledge   &88.0       &59.6   &73.3             \\

BloombergGPT       &Static Knowledge   &51.1       &75.1   &-             \\

Llama2-7B      &General   &39.0       &80.0   &29.6             \\

FLANG      &Static Knowledge   &91.9       &3.4   &-             \\
\bottomrule
\end{tabular}%
}
\caption{ Performance comparison of representative models under the four knowledge injection paradigms on financial benchmarks.}
\label{compare_finance}
\end{table}


To systematically compare the effectiveness of different knowledge injection paradigms in practical settings, we focus on two representative domains: biomedicine and finance. 

In the biomedical domain, which is widely studied and rich in benchmark datasets (e.g., MedQA, PubMedQA, and MedMCQA), we evaluate the performance of various models (see Table~\ref{compare}). Although the models differ in architecture, we align their backbones where possible—for example, both PMC-LLaMA and MedAdapter use LLaMA-13B. For closed-source models like GPT-4, prompt engineering is the only feasible adaptation strategy. Nevertheless, GPT-4 combined with Medprompt achieves strong performance, demonstrating the effectiveness of prompt-based knowledge injection. Among open models, MedAdapter underperforms compared to fully fine-tuned models such as PMC-LLaMA, suggesting that full fine-tuning may be more effective than adapter-based methods for certain tasks. Models with static knowledge (e.g., MedBERT) show substantial variance across tasks, underscoring the importance of alignment between pretraining corpora and downstream objectives.

In Table \ref{compare_finance}, we extend this comparison to the financial domain, evaluating models on benchmarks such as FPB\cite{Malo2014GoodDO}, FiQA-SA\cite{maia201818}, and TFNS\cite{el-haj-etal-2020-financial}. 
The findings closely mirror those in the biomedical domain. Finetuned GPT-4 consistently outperforms others, confirming the value of injecting domain-specific knowledge into general-purpose LLMs. Static knowledge models like FinBERT and FLANG perform well on certain tasks but show significant variability, again emphasizing the crucial role of corpus-task alignment. FinGPT, which adopts lightweight adapter-based knowledge injection, achieves competitive performance while maintaining adaptability. In contrast, LLaMA2-7B lags behind across most tasks, reinforcing the necessity of targeted knowledge injection for domain-intensive applications. The consistency of observations across both domains suggests that the effectiveness of knowledge injection depends on a careful balance of architectural design, adaptation strategy, and corpus alignment to support complex, high-stakes tasks.

\section{Detailed Discussions on Challenges \& Opportunities}
\label{challenges}

\subsection{Integrated Knowledge Consistency}
While knowledge injection empowers LLMs to reason with external facts, it introduces a crucial consistency problem: injected knowledge may contradict either the model’s internal representations or other pieces of retrieved information~\cite{xu2024knowledge}. 
In high-stakes domains such as healthcare and law, even minor inconsistencies can lead to significant consequences—for instance, conflicting drug dosages from different clinical guidelines~\citep{dayton2012standards,zhao2025medrag}, or divergent legal interpretations across jurisdictions~\citep{guha2023legalbench}.

Recent research proposes techniques such as post-retrieval contradiction detection \citep{xu2024sparsecl}, and confidence-aware re-ranking \citep{ren2025self} to address these issues. 
Some frameworks, like MedRAG\citep{zhao2025medrag}, apply weighted retrieval and ensemble voting to prioritize reliable sources.
Others explore neural symbolic consistency checking, where injected knowledge is aligned to pre-defined ontologies or verified using structured reasoning paths~\citep{ciatto2024symbolic}.

Another growing area involves alignment-aware reranking, where retrieved documents are filtered based on their alignment with the LLM’s intermediate beliefs~\citep{jin2025llm}. 
Future directions may include interactive consistency resolution (e.g., user-in-the-loop conflict selection), as well as integrating factual calibration modules~\citep{dong2022calibrating} that explicitly monitor factuality during decoding. These methods collectively aim to make knowledge-enhanced LLMs more robust, explainable, and reliable in dynamic or sensitive environments.

\subsection{Cross-Domain Knowledge Transfer}
Cross-domain transfer is a central challenge in building generalized yet specialized LLMs. As LLMs are exposed to knowledge from diverse domains, they must navigate incompatible ontologies, varied domain languages, and distinct reasoning structures~\cite{li2025zero}. For instance, transferring concepts from chemistry to healthcare involves not only bridging terminology gaps but also adapting to different causal assumptions and data formats~\citep{schroeder2018relating}.

Several strategies have been proposed to manage this complexity. 
Adapter-based modularization~\cite{he2021effectiveness} allows domain-specific components to be trained separately and selectively activated. 
Meta-learning approaches\citep{hou2022meta} help models rapidly adapt to new domains with minimal supervision. Additionally, continual pretraining on mixed-domain corpora~\citep{jin2022lifelong} offers a scalable method to improve robustness without catastrophic forgetting.

Standardized datasets such as CrossNER\citep{liu2021crossner} for multilingual named entity recognition, MultiLexSum\citep{shen2022multi} for cross-domain summarization, and MEDIQA-QA~\citep{yadav2021nlm} for biomedical QA serve as valuable testbeds for cross-domain evaluation. Future work may explore retrieval-augmented transfer, where dynamic selection of domain-relevant knowledge supports adaptive reasoning, or domain-invariant embedding learning, enabling LLMs to generalize across tasks without explicit supervision.

\subsection{Scalability and Efficiency of Knowledge Integration}

As LLMs are increasingly augmented with large-scale external knowledge such as entire knowledge graphs, document corpora, or real-time retrieval APIs, the computational and memory cost of incorporating this knowledge becomes a bottleneck.
Efficient integration remains a key challenge, especially when operating under low-resource or latency-constrained settings.

Techniques such as sparse retrieval~\citep{lee2019latent}, memory compression~\citep{zhong2024memorybank}, and caching strategies have been proposed to reduce overhead. Modular architectures (e.g., adapters or plug-in modules) also allow partial activation of knowledge, improving scalability. Future research could explore task-aware pruning of external knowledge, knowledge distillation from retrieval-based pipelines into compact models, and efficient routing mechanisms to select only relevant knowledge for each input.

\subsection{Evaluation and Hallucination Detection}

Evaluating knowledge-enhanced LLMs remains difficult due to the lack of standardized benchmarks and automatic metrics for factual consistency, coverage, and reasoning depth. Moreover, LLMs often hallucinate facts even when augmented with accurate knowledge~\citep{ji2023survey}, making it hard to trust outputs in high-stakes tasks.

Recent work explores metrics like FactScore \citep{min2023factscore}, entailment-based verification \citep{patwa2022benchmarking}, and human-in-the-loop evaluation schemes. However, few of these methods scale across domains or languages. There is a growing need for task-specific, fine-grained evaluation metrics that capture whether the model used the retrieved knowledge effectively and truthfully. 
Additionally, incorporating hallucination detection as an internal module, such as through consistency checks between generation and source knowledge, may help reduce risk and improve interoperability.

\end{document}